\documentclass[lettersize,journal]{IEEEtran}
\IEEEoverridecommandlockouts
\usepackage{amsthm}

\usepackage{times,amsmath,epsfig}
\usepackage{epstopdf}
\usepackage{amssymb}
\usepackage{scalerel}
\usepackage{cite}
\usepackage{caption}	
\usepackage{graphicx}
\usepackage{url}
\usepackage[bookmarks=false]{hyperref}
\usepackage{multirow}
\usepackage{multicol}
\usepackage{subfigure}
\usepackage{threeparttable}
\usepackage{booktabs}
\usepackage{amsfonts}
\usepackage{textcomp}
\usepackage[framemethod=1]{mdframed}
\usepackage{enumitem}
\usepackage{enumerate}

\newcommand{\tabincell}[2]{\begin{tabular}{@{}#1@{}}#2\end{tabular}}
\ifodd 1
    
    \newcommand{\com}[1]{\textbf{\color{red} (COMMENT: #1)}} %comment of the text
\else
    
    \newcommand{\com}[1]{}
\fi
\title{\huge \textsf{DeepOPF-AL}: Augmented Learning for Solving AC-OPF Problems with Multiple Load-Solution Mappings}

\author{Xiang Pan, Wanjun Huang, Minghua Chen, and Steven H. Low \thanks{X. Pan and W. Huang are with The Chinese University of Hong Kong. M. Chen is with City University of Hong Kong. S. Low is
with California Institute of Technology. Corresponding author: Minghua Chen.} 
}

\begin{document}
\maketitle

\begin{abstract} 
The existence of multiple load-solution mappings of non-convex AC-OPF problems poses a fundamental challenge to deep neural network (DNN) schemes. As the training dataset may contain a mixture of data points corresponding to different load-solution mappings, the DNN can fail to learn a legitimate mapping and generate inferior solutions. We propose \textsf{DeepOPF-AL} as an \textit{augmented-learning} approach to tackle this issue. The idea is to train a DNN to learn a \textit{unique} mapping from an augmented input, i.e., (load, initial point), to the solution generated by an iterative OPF solver with the load and initial point as intake. We then apply the learned augmented mapping to solve AC-OPF problems much faster than conventional solvers. Simulation results over IEEE test cases show that \textsf{DeepOPF-AL} achieves noticeably better optimality and similar feasibility and speedup performance, as compared to a recent DNN scheme, with the same DNN size yet elevated training complexity.

%% The results also show the robust performance of \textsf{DeepOPF-AL}.

\end{abstract}

\begin{IEEEkeywords}
optimal power flow; augmented learning 
\end{IEEEkeywords}
% deep neural network

%% hwj for reference
%hwj: The idea is to train a DNN to learn a \textit{unique} mapping from an augmented input, \textcolor{green}{i.e., load and initial point,} to the...feasibility and speedup \textcolor{green}{performances, as compared to the state-of-the-art scheme}.

\section{Introduction}\label{introduction}
Machine learning has recently been employed to solve optimal power flow (OPF) problems efficiently. By applying supervised learning techniques and using prepared (load, optimal solution) data,~\cite{ng2018statistical,zhang2021convex} indicate {deep neural network (DNN)} could reduce the computation time of the conventional method by determining active constraints upon the given load. The studies in \cite{deepopf1} and later in~\cite{deepopfv,fioretto2020predicting,zamzam2020learning} train DNN for solving OPF problems by learning the high-dimensional load-solution mapping, which could directly generate feasible and close-to-optimal solutions much faster than conventional solvers.

%Recently, learning techniques like deep neural network (DNN) provide a promising avenue to solve optimal power flow (OPF) problems. By applying supervised learning framework and using prepared (load, optimal solution) data, the studies in \cite{deepopf1} and later in ~\cite{deepopfv,fioretto2020predicting,zamzam2020learning} train DNN for solving OPF problems by learning the high-dimensional load-solution mapping and then using it to directly generate feasible and close-to-optimal solutions much faster than conventional solvers.

%Existing works apply supervised learning frameworks to train DNN using the obtained solution of pre-solved load input. 

While existing works suggest the potential of DNN in solving AC-OPF problems, the presence of multiple load-solution mappings poses a fundamental challenge to the supervised learning-based schemes. As the non-convex AC-OPF problem may admit multiple optimal solutions for a load input~\cite{bukhsh2013local}, the training dataset may contain ``mixed'' data points, i.e., the solutions to different load inputs correspond to different mappings, making the learning task inherently difficult. Consequently, the trained DNN can fail to learn a legitimate mapping (one of the multiple target mappings) and generate inferior solutions~\cite{kotary2021learning}. See Fig.~\ref{fig:2bus.example} for an illustrating example. Since the solutions to a load from different mappings can derive similar objective values or exhibit similar solution structures (e.g., both solutions have ``high-voltage'' solutions on the same buses and ``low-voltage'' solutions on the remaining buses), it is non-trivial to differentiate them from each other. Consequently, intuitive methods such as selecting the least-cost and ``high-voltage'' solution for each load may fail to address the issue.
%deepopf3
%One may consider addressing this issue by providing one solution (e.g., the least-cost or high-voltage solution) for each load input. Still, the selected data points can correspond to different mappings, and hence it cannot solve the issue of having mixed data points from multiple mappings.} 

There exist two approaches to tackle the challenge. One is to prepare the training dataset so that it contains data points from only one mapping~\cite{kotary2021learning}. The other is to apply unsupervised learning to train DNN without labeling data points~\cite{deepopfngt}. Yet, their limitations lie in (i) substantial computational complexity in preparing training data and in unsupervised learning and (ii) no guarantee to learn a legitimate mapping for AC-OPF problems with multiple load-solution correspondences.

In this paper, we propose \textsf{DeepOPF-AL} as an augmented learning approach to learn a \textit{unique} mapping from an augmented load input to the corresponding solution, and then use it to solve AC-OPF problems. Our contribution is two-fold. First, after presenting a simple example showing AC-OPF problems can indeed admit multiple load-solution mappings, we develop \textsf{DeepOPF-AL} following a particular augmented-learning design. Specifically, we train a DNN to learn the mapping from (load, initial point) to the unique OPF solution generated by the Newton-Raphson method with the load and initial point as intake\footnote{We note that \textsf{DeepOPF-AL} is different from the approach in~\cite{baker2020learning} in that it directly outputs the solution in one pass while the latter is an iterative scheme that replaces the update function in the Newton-Raphson method by a DNN.}. Second, simulation results on IEEE test cases show that \textsf{DeepOPF-AL} achieves better AC-OPF optimality and similar feasibility and speedup performance, as compared to a recent scheme~\cite{deepopfv}, with the same DNN size yet elevated training complexity. Simulation results also show robust performance of \textsf{DeepOPF-AL}.
\section{AC-OPF and Multiple Load-solution Mappings}\label{sec:acopf.multiple.solution.mapping}
\begin{figure}[!t]
	\centering
	%\subfigure [] {\includegraphics[width = 0.22\textwidth]{fig/figure1_1}}
	\subfigure [] {\includegraphics[width = 0.15\textwidth]{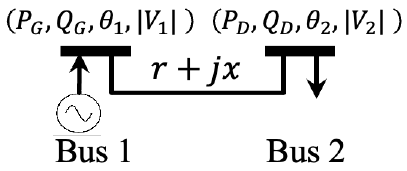}}
    %  \hfill
	\subfigure [] {\includegraphics[width = 0.32\textwidth]{fig/2.bus.mapping.v6}}	
	\captionsetup{font={small}}
	\caption{(a) A 2-bus power network with $|V_1| = 0.9$ p.u., $\theta_1=0$, $r = 0$ p.u., $x = 0.25$ p.u., and $P_D = 343$ MW. (b) The AC-OPF problem in \eqref{obj}-\eqref{acopf_con6} over the 2-bus network has two legitimate mappings from the load $Q_D$ to $|V^*_2|$ in the optimal solution. A DNN trained with data uniformly sampled from all possible $(Q_D, |V^*_2|)$ pairs and using $\ell_2$ distance as the loss function fails to learn any of the two mappings.}
	\label{fig:2bus.example}
\end{figure}

The standard AC-OPF problem is formulated as
%based on the bus-injection model 
{\small
\begin{align}
    {\min}&  \sum\nolimits_{i \in \mathcal{N}}{C_i\left( P_{Gi} \right)} \label{obj}\\
    % \label{acopf_con1}
    \mathrm{s.t.}&
    \sum\nolimits_{\left( i,j \right) \in \mathcal{E}}{\text{Re}\left\{ V_i\left( V_{i}^{*}-V_{j}^{*} \right) y_{ij}^{*} \right\}}=P_{Gi}-P_{Di},  i \in \mathcal{N}, \label{acopf_con2} \\
    &\sum\nolimits_{\left( i,j \right) \in \mathcal{E}}{\text{Im}\left\{ V_i\left( V_{i}^{*}-V_{j}^{*} \right) y_{ij}^{*} \right\}}=Q_{Gi}-Q_{Di}, i\in \mathcal{N}, \label{acopf_con3} \\
    &P_{Gi}^{\min}\le P_{Gi}\le P_{Gi}^{\max}, i\in \mathcal{N}, \label{acopf_con4} \\
    &Q_{Gi}^{\min}\le Q_{Gi}\le Q_{Gi}^{\max}, i\in \mathcal{N},\label{acopf_con5}\\
    &V_{i}^{\min}\le |V_i| \le V_{i}^{\max}, i\in \mathcal{N},\label{acopf_con6}\\
    &| V_i\left( V_{i}^{*}-V_{j}^{*} \right) y_{ij}^{*} |\le S_{ij}^{\max},\left( i,j \right) \in \mathcal{E},\label{acopf_con7}\\
    \mathrm{var.}&\;\; P_{Gi},Q_{Gi}, V_i, i\in \mathcal{N}. \nonumber
\end{align}}
\noindent $\text{Re} \{z\}$, $\text{Im}\{z\}$, $z^*$, and $|z|$ denote the real part, the imaginary part,  the conjugate, and the magnitude of a complex variable $z$, respectively. $\mathcal{N}$ and $\mathcal{E}$ denote the set of buses and the set of edges, respectively. $P_{Gi}$ (resp. $Q_{Gi}$) and $P_{Di}$ (resp. $Q_{Di}$) denote the active (resp. reactive) power generation and active (resp. reactive) load on bus $i$, respectively. $V_i$ represents complex voltage, including the magnitude $|V_i|$ and the phase angle $\theta_i$, on bus $i$. %$\theta_{ij}$ is the difference between $\theta_i$ and $\theta_j$.
$y_{ij}$ and $S_{ij}^{\max}$ denote the admittance and the branch flow limit of the branch $(i,j) \in \mathcal{E}$, respectively. \eqref{acopf_con2} and \eqref{acopf_con3} represent power-flow balance equations. \eqref{acopf_con4} and \eqref{acopf_con5} represent the active and reactive generation limits. \eqref{acopf_con6} represents the voltage magnitude limit, and \eqref{acopf_con7} represents the branch flows limits. The objective is to
minimize the total cost of active power generation, where $C_{i}(\cdot)$ is the quadratic cost function of the generator at bus $i$. We set $C_{i}(P_{Gi})=0$ and $P_{Gi}^{\min}=P_{Gi}^{\max}=Q_{Gi}^{\min}=Q_{Gi}^{\max}=0$ if bus $i$ has no generator. 
As shown in Fig.~\ref{fig:2bus.example}, existing learning-based schemes fail to learn either of two legitimate mappings. 

%and generate satisfied solution.
%with inferior optimality performance. %See Fig.~\ref{fig:2bus.example} for an example.in Sec.~\ref{introduction}
%reduce the following original paragraph to above sentences
%As discussed above, existing DNN schemes learn a load-solution mapping of the AC-OPF problem and use the learned mapping to generate solutions efficiently. However, as the AC-OPF problem is non-convex and may admit multiple optimal solutions for a load input, the training dataset may contain a mixture of data points corresponding to multiple load-solution mappings. Consequently, existing DNN schemes can fail to learn a legitimate mapping and generate solutions with inferior optimality performance. See Fig.~\ref{fig:2bus.example} for an example.

\section{DeepOPF-AL: Solving AC-OPF Problems by Learning Unique Augmented Mapping}\label{deepopfa.framework}
\begin{figure}[!t]
	\centering
	\includegraphics[width = 0.5\textwidth]{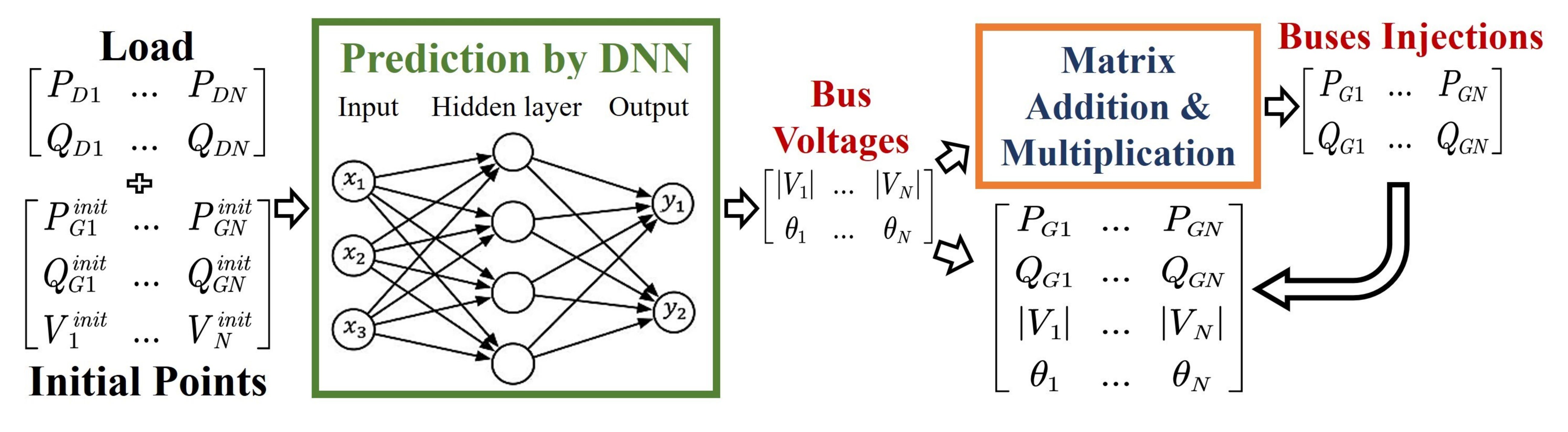}
    \captionsetup{font={small}}
	\caption{Illustration of the \textsf{DeepOPF-AL} approach. Given a load input and an initial point, we employ a DNN model to predict the bus voltages and reconstruct the remaining bus injection and generation solutions by simple scalar computations using power flow equations.
	} %generated by a deterministic iterative algorithm, with the load and initial point as intake
	\label{fig:DeepOPF-AL.illustration}
\end{figure}

The schematic of the proposed \textsf{DeepOPF-AL} is shown in Fig.~\ref{fig:DeepOPF-AL.illustration}. It follows a particular augmented-learning design to train a DNN to learn the mapping from (load, initial point) to the unique AC-OPF solution generated by the Newton-Raphson method with the load and initial point as intake. We build the DNN model on the multi-layer feed-forward neural network structure with ReLU as the activation function of hidden layers. We design the loss function as the total mean square error between the generated solution and the ground truth (the generated AC-OPF solution). We apply the popular Adam algorithm to update the DNN's parameters in training.
%The training load data is sampled using the real-world demand curve and the initial points are sampled uniformly at random in the feasible set.
%\com{briefly describe the NN structure, activation function, output layer sigmoid?, the training data generation, the loss function, and the training algorithm.}
\textsf{DeepOPF-AL} uses the trained DNN to predict the bus voltages and reconstructs the bus injections, i.e., RHS values of~\eqref{acopf_con2}-\eqref{acopf_con3}, and finally the generations, all by simple scalar calculation. Such a predict-and-reconstruct framework~\cite{deepopf1,deepopf2,deepopfv} guarantees the power-flow equality constraints and reduces the number of variables to predict. Lastly, \textsf{DeepOPF-AL} employs the post-processing process in~\cite{deepopfv} to help keep the obtained solution within the box constraints in~\eqref{acopf_con4}-\eqref{acopf_con6}. 

We now explain the \textit{unique} augmented mapping learned by \textsf{DeepOPF-AL}. Denote $\mathcal{L}(\mathbf{Z})$ as the Lagrangian of the AC-OPF problem in \eqref{obj}-\eqref{acopf_con7}, where $\mathbf{Z}$ is the concatenation of the primal variables, the dual variables, the slack variables, and the load input {(denoted as $\mathbf{D}$)}. Let $\mathbf{Z}_0$ and $\mathbf{Z}_t$ denote the initial values of $\mathbf{Z}$, and the updated ones by the Newton-Raphson method after the $t$-th iteration, respectively. Similarly, let $\mathbf{X}_0:=( P^{\mbox{init}}_{Gi},Q^{\mbox{init}}_{Gi}, V^{\mbox{init}}_{i},i\in \mathcal{N})$ and $\mathbf{X}_t$ denote the initial conditions for the primal variables and the updated ones after the $t$-th iteration, respectively. 
Recall that the Newton-Raphson method works as follows: for the $t$-th iteration, it first computes the gradient and Hessian of the Lagrangian with respect to $\mathbf{Z}_t$, i.e., $\nabla \mathcal{L}(\mathbf{Z}_t)$ and ${\nabla}^2 \mathcal{L}(\mathbf{Z}_t)$. It then computes the update step for $\mathbf{Z}_t$ as $\varDelta \mathbf{Z}_t=-\nabla \mathcal{L}(\mathbf{Z}_t)\cdot \left({\nabla}^2 \mathcal{L}(\mathbf{Z}_t)\right)^{-1}$. It extracts  $\varDelta \mathbf{X}_t$ from $\varDelta \mathbf{Z}_t$ and  generates $\mathbf{X}_{t+1}=\mathbf{X}_t+\alpha_t \varDelta \mathbf{X}_t$ with a pre-determined step-size $\alpha_t$. Hence, given $\mathbf{X}_0$ and $\mathbf{D}$, $\mathbf{X}_1$ and all follow-up $\mathbf{X}_t$s are uniquely determined. Finally, the iterations terminate when pre-specified termination criteria are satisfied. Both the total number of iterations and final solution are unique for given initial point $\mathbf{X}_0$ and load $\mathbf{D}$. Overall, we observe a unique mapping from ($\mathbf{X}_0, \mathbf{D}$) to the corresponding final solution, described as $\mathbf{X}_{\mbox{NR}}\left(\mathbf{X}_0, \mathbf{D}\right)=\psi_{\mbox{NR}}\left( \mathbf{X}_0,\mathbf{D} \right)$. \textsf{DeepOPF-AL} aims to learn this unique mapping $\psi_{\mbox{NR}} \left(\cdot \right)$.

%\textsf{DeepOPF-AL} learns the composite mapping $\phi _{T(\mathbf{X}_0,\mathbf{D})}\left(\cdots\left( \phi _0\left( \mathbf{X}_0, \mathbf{D} \right) \right)\right)$. }%\rev{As such, \textsf{DeepOPF-AL} resolves the issue of having mixed data points and guarantees to learn a \textit{legitimate} mapping.}
% \begin{eqnarray}
%     f_{\mbox{aug}}^{*}\left( \mathbf{X}_0, \mathbf{D} \right) \triangleq \phi _T\left(\cdots\phi _2\left( \phi _1\left( \mathbf{X}_0, \mathbf{D} \right) \right)\right).  \label{mapping}
% \end{eqnarray}

\subsubsection*{Discussion} (i) \textsf{DeepOPF-AL} can learn the augmented mapping of any deterministic iterative algorithm. We choose to learn that of the popular Newton-Raphson method. {For a load input, one can run several \textsf{DeepOPF-AL} in parallel with different initial points and output the least-cost solution.} (ii) \textsf{DeepOPF-AL} learns a unique augmented mapping, but requiring a larger DNN and more training data than learning a standard load-solution one, as observed in Sec.~\ref{sec:simulation.results}. (iii) As we will also see in 
Sec.~\ref{sec:simulation.results}, for the set of (load, initial point) inputs for which the Newton-Raphson method fails to converge, \textsf{DeepOPF-AL} can still generate solutions with decent optimality performance. This indicates better practicability of DNN schemes over iterative solvers, in addition to speedup.

% While it is expected to generate close-to-optimal solutions, there may exist initial points for which \textsf{DeepOPF-AL} gives sub-optimal solutions with notable cost differences. In practice, for a load input, one can generate several initial points and run several \textsf{DeepOPF-AL} in parallel to obtain multiple AC-OPF solutions, and output the least-cost one.

%In our implementation and simulations in Sec.~\ref{sec:simulation.results}, for each load input, we randomly generate several initial points and run several \textsf{DeepOPF-AL} in parallel to generate multiple AC-OPF solutions. We output the one with lowest cost and record the worst running time. }
%experimental items
%1. case with multiple solutions, e.g., case39, to show the large cost difference between DeepOPF-AL and DeepOPF-V
%2. show DeepOPF-AL can work for large-scale system, e.g., case118 and case2000
\section{Numerical Experiments}\label{sec:simulation.results}
We conduct simulations in CentOS 7.6 with quad-core (i7-3770@3.40G Hz) CPU and 16GB RAM. We compare performance of \textsf{DeepOPF-AL} and a recent scheme \textsf{DeepOPF-V}~\cite{deepopfv} on IEEE 39-/300-bus systems. We generate realistic load on each bus by multiplying the default value by an interpolated demand curve based on 11-hour California’s net load in Jul. -- Sept. 2021, with a time granularity of 30 seconds thus 2,760 load instances per day. For each load, we randomly generate initial points and fed them together with the loads into the Matpower Interior Point Solver (MIPS) solver~\cite{zimmerman2011matpower}, which implements the Newton-Raphson method, to obtain reference AC-OPF solutions. Our DNN models consist of 3 hidden layers with 1024/768/512 neurons for the 39-bus system and 4 hidden layers with 1024/768/512/256 neurons for the 300-bus system. We set the batch size, maximum epoch, and learning rate to be 50, 4000, and 1e-4, respectively. 

\subsubsection*{Performance metrics} (i) the relative optimality difference, i.e., $\eta_{opt}$, between the objective values obtained by \textsf{DeepOPF-AL} and MIPS; (ii) the average running times of the MIPS solver, i.e., $t_{mips}$, the DNN schemes, i.e., $t_{dnn}$, and the corresponding average speedup ratios, i.e., $\eta_{sp}$; (iii) the average constraint satisfaction percentages for active/reactive generation and branch flow limit, i.e., $\eta_{P_{G}}, \eta_{Q_{G}}$, and $\eta_{S_{l}}$, respectively; (iv) the average load-serving mismatch percentage of active and reactive loads, i.e., $\eta_{P_D}$ and $\eta_{Q_D}$, respectively.

%unbalance with 49000 data
% \begin{table}[t]
% % \normalsize
% \centering
% \caption{Simulation results for the IEEE 39-bus system.}
% \renewcommand\arraystretch{1.0}
% \begin{threeparttable}
% \scalebox{1.0}{
% \begin{tabular}{c|c|c|c|c}		
% \hline 
% \multirow{3}{*}{Metric}&
% \multicolumn{2}{c|}{\tabincell{c}{Balanced Dataset}} &
% \multicolumn{2}{c}{\tabincell{c}{Unbalanced Dataset}}\\
% \cline{2-5}
% \multicolumn{1}{c|}{} & \tabincell{c}{\textsf{DeepOPF}\\-AL} & \tabincell{c}{\textsf{DeepOPF-V}}&\tabincell{c}{\textsf{DeepOPF}\\-AL} & \tabincell{c}{\textsf{DeepOPF-V}}\\
% \hline 
% $\eta_{opt}$(\%) &-0.74 &-13.5 & -1.78& -7.21\\
% $\eta_{\boldsymbol{P_G}}$(\%) &97.8&99.3&99.0&99.8 \\
% $\eta_{\boldsymbol{Q_G}}$(\%) &96.1&92.6&97.4&96.4 \\
% $\eta_{\boldsymbol{S_l}}$(\%) &99.8&100&99.9&100 \\
% $\eta_{\boldsymbol{P_D}}$(\%) &0.39&0.56&0.21&0.27 \\
% $\eta_{\boldsymbol{Q_D}}$(\%) &3.86&24.2&2.19&11.2 \\
% $t_{mips}$ (ms)&1621&1621&1987&1987 \\
% $t_{dnn}$ (ms)&1.4&1.3&1.3&1.2 \\
% $\eta_{sp}$ &$\times$1157&$\times$1247&$\times$1528&$\times$1655\\
% %\hline
% %$\eta_{sp}$ &\multicolumn{2}{c|}{$\times$1080}&\multicolumn{2}{c}{$\times$1528}\\
% \hline  			
% \end{tabular}}
% \end{threeparttable}
% \label{table1}
% \end{table}
%unbalance with 49000 data
%the results with smallest cost
\begin{table}[t]
% \normalsize
\centering
\caption{Simulation results for the IEEE 39-bus system.}
\renewcommand\arraystretch{1.0}
\begin{threeparttable}
\scalebox{1}{
\begin{tabular}{c|c|c|c|c}		
\hline 
\multirow{3}{*}{Metric}&
\multicolumn{2}{c|}{\tabincell{c}{Balanced Dataset}} &
\multicolumn{2}{c}{\tabincell{c}{Unbalanced Dataset}}\\
\cline{2-5}
\multicolumn{1}{c|}{} & \tabincell{c}{\textsf{DeepOPF}\\-AL} & \tabincell{c}{\textsf{DeepOPF-V}}&\tabincell{c}{\textsf{DeepOPF}\\-AL} & \tabincell{c}{\textsf{DeepOPF-V}}\\
\hline 
$\eta_{opt}$(\%) &0.48 &-8.56 & 0.66& -5.98\\
$\eta_{\boldsymbol{P_G}}$(\%) &97.5&99.3&97.4&99.8 \\
$\eta_{\boldsymbol{Q_G}}$(\%) &94.6&91.7&98.4&96.7 \\
$\eta_{\boldsymbol{S_l}}$(\%) &99.9&100&99.9&100 \\
$\eta_{\boldsymbol{P_D}}$(\%) &0.19&0.61&0.09&0.32 \\
$\eta_{\boldsymbol{Q_D}}$(\%) &6.47&27.6&0.49&12.9 \\
$t_{mips}$ (ms)&1621&1621&1987&1987 \\
$t_{dnn}$ (ms)&1.4&1.3&1.3&1.2 \\
$\eta_{sp}$ &$\times$1157&$\times$1247&$\times$1528&$\times$1655\\
%\hline
%$\eta_{sp}$ &\multicolumn{2}{c|}{$\times$1080}&\multicolumn{2}{c}{$\times$1528}\\
\hline  			
\end{tabular}}
\end{threeparttable}
\label{table1}
\end{table}

\begin{table}[t]
    \centering
    \caption{Simulation results for the IEEE 300-bus system.}
    \renewcommand\arraystretch{1}
    \centering
    %\normalsize
    \begin{threeparttable}
         \scalebox{1}{
        \begin{tabular}{c|c|c|c|c}		
            \hline 
            \multirow{2}{*}{Metric}&
            \multicolumn{2}{c|}{\tabincell{c}{Convergent Dataset}} &
            \multicolumn{2}{c}{\tabincell{c}{Non-convergent Dataset}} \\
            \cline{2-5}
            \multicolumn{1}{c|}{} & \tabincell{c}{\textsf{DeepOPF-AL}} & \tabincell{c}{\textsf{DeepOPF}-V}&\tabincell{c}{\textsf{DeepOPF-AL}} & \tabincell{c}{MIPS} \\
            \hline 
            $\eta_{opt}$(\%) &0.01&-0.01&-0.13&20.4 \\
            $\eta_{\boldsymbol{P_G}}$(\%)&100&100&99.9&23.2 \\
            $\eta_{\boldsymbol{Q_G}}$(\%) &100&100&100&78.3 \\
            $\eta_{\boldsymbol{S_l}}$(\%) &100&100&100&80.7\\
            $\eta_{\boldsymbol{P_D}}$(\%) &0.0&-0.01&-0.1&-73.4\\
            $\eta_{\boldsymbol{Q_D}}$(\%) &0.02&0.02&-0.02&-109.2\\
            $t_{mips}$(ms)&2978&2978&-&- \\
            $t_{dnn}$(ms)&2.2&1.8&2.2&- \\
            $\eta_{sp}$&$\times$1353&$\times$1654&-&-\\
            \hline  			
        \end{tabular} }
    \end{threeparttable}
    \label{table2}
\end{table}

\subsubsection*{The case with multiple load-solution mappings} 
We evaluate the performance of \textsf{DeepOPF-AL} and \textsf{DeepOPF-V} with the same DNN size and training/testing data (for fair comparisons) on an IEEE 39-bus system having two load-solution mappings with on average 30\% difference in objective values~\cite{bukhsh2013local}. We design two datasets: balanced dataset (with 89,972 data points; for each load, the ratio between the numbers of two solutions is 1:1) and unbalanced dataset (with 52,384 data points; for each load, the ratio between the numbers of the low-cost solutions and the high-cost solutions are 9:1). We split each dataset with the ``80/20'' strategy to obtain the training set and test set, respectively. The results are shown in Table~\ref{table1}. As seen, \textsf{DeepOPF-AL} outperforms \textsf{DeepOPF-V} in both datasets and achieves smaller optimality gap and {similar} constraint and load satisfaction percentages.  \textsf{DeepOPF-AL} performs more consistently than \textsf{DeepOPF-V} over the two datasets, demonstrating its effectiveness in solving AC-OPF problems with multiple load-solution mappings.

\subsubsection*{The case with unique load-solution mapping and robust performance}
We evaluate the performance of \textsf{DeepOPF-AL} and \textsf{DeepOPF-V} on the IEEE 300-bus test system, which we observe has a unique load-solution mapping. We use the same size DNN but different size dataset, following the ``80/20'' training/testing splitting rule. The training set for \textsf{DeepOPF-V} contains 2,760 (load, solution) data pairs. The training set for \textsf{DeepOPF-AL} has 110,400 ((load, initial point), solution) data points, where we randomly sample 40 initial points for each load. The results in Table~\ref{table2} show that \textsf{DeepOPF-AL} achieves similar optimality gap, feasibility, and speedup performance, as compared to \textsf{DeepOPF-V}. These results confirm that, for this setting with a unique load-solution mapping, the (higher-dimensional) augmented mapping to learn by \textsf{DeepOPF-AL} is degenerated and can be represented using the same-size DNN as the (lower-dimensional) load-solution one to learn by \textsf{DeepOPF-V}. Meanwhile, the augmented mapping still requires more data to train, if one has no prior knowledge of its degenerated dimension, as in this simulation.

We also test \textsf{DeepOPF-AL} using 2760$\times$20 (load, initial point) data for which the MIPS solver fails to converge. We observe in Table~\ref{table2} that it still attains decent performance, while the non-convergent solutions by the MIPS solver suffer from significant performance degradation. This implies that \textsf{DeepOPF-AL}, while trained with data generated by the MIPS solver, achieves more robust performance than the solver.

\section{Concluding Remark}
We propose \textsf{DeepOPF-AL} as the first learning-based approach that guarantees to learn a unique augmented mapping for solving AC-OPF problems that potentially admit multiple load-solution correspondences. Simulation results show that it achieves a smaller optimality gap and similar feasibility and speedup performances compared to a recent DNN scheme, with elevated training complexity. A future direction is to improve augmented-learning designs for better training efficiency for solving AC-OPF and other non-convex problems.

%\rev{A future direction is to explore alternative augmented-learning designs to improve the training efficiency of \textsf{DeepOPF-AL} for solving AC-OPF problems and general non-convex optimization problems.} We believe augmented learning can also find applications in mapping approximation in other domains.

%the proposed augmented learning approach shares similar spirits as the kernel method in classification~\cite{shawe2004kernel}. The differences lie in the application (regression vs. classification) and purpose (learning a mapping out of multiple ones vs. learning one difficult mapping). 

%\rev{Improving training efficiency of augmented learning schemes is an interesting future direction.} 
% without sacrificing performance 
% We also believe that augmented learning can find application in general mapping approximation in other problem domains.  

\bibliographystyle{IEEEtran}
\bibliography{IEEEabrv,ref}
\end{document}